\documentclass{IEEEtran4PSCC}

\usepackage[cmex10]{amsmath}
\usepackage{amsfonts,amssymb,amsthm}
\usepackage{pifont}
\usepackage{multirow}

\ifCLASSINFOpdf
   \usepackage[pdftex]{graphicx}
\else
   \usepackage[dvips]{graphicx}
\fi
\usepackage{float}
\ifCLASSOPTIONcompsoc
 \usepackage[caption=false,font=normalsize,labelfont=sf,textfont=sf]{subfig}
\else
 \usepackage[caption=false,font=footnotesize]{subfig}
\fi

\floatstyle{ruled}
\newfloat{model}{thp}{lop}
\floatname{model}{Model}

\newtheorem{theorem}{Theorem}
\newtheorem{lemma}{Lemma}
\usepackage{algorithmic}
\usepackage[ruled,linesnumbered,noresetcount,vlined]{algorithm2e}

\usepackage{url}

\usepackage[dvipsnames]{xcolor}
\usepackage{booktabs}
\usepackage{todonotes}
\usepackage{accents}
\usepackage{nicefrac}
\usepackage{cite}

\newcommand{\dcp}{{\sc DCP}}
\definecolor{revColor}{rgb}{0,0,0}
\newcommand{\revision}[2][revColor]{\textcolor{#1}{#2}}

\makeatletter
\DeclareRobustCommand{\cev}[1]{%
  {\mathpalette\do@cev{#1}}%
}
\newcommand{\do@cev}[2]{%
  \vbox{\offinterlineskip
    \sbox\z@{$\m@th#1 x$}%
    \ialign{##\cr
      \hidewidth\reflectbox{$\m@th#1\vec{}\mkern4mu$}\hidewidth\cr
      \noalign{\kern-\ht\z@}
      $\m@th#1#2$\cr
    }%
  }%
}

\newcommand{\ubar}[1]{\underaccent{\bar}{#1}}

\makeatother

\newcommand{\im}{\mathbf{j}}

\newcommand{\pg}{\mathbf{p}^{\text{g}}}
\newcommand{\qg}{\mathbf{q}^{\text{g}}}
\newcommand{\Sg}{\mathbf{S}^{\text{g}}}

\newcommand{\vm}{\mathbf{v}}
\newcommand{\va}{\mathbf{\theta}}
\newcommand{\V}{\mathbf{V}}

\newcommand{\pf}{\vec{\mathbf{p}}}
\newcommand{\qf}{\vec{\mathbf{q}}}
\newcommand{\Sf}{\vec{\mathbf{S}}}
\newcommand{\pt}{\cev{\mathbf{p}}}
\newcommand{\qt}{\cev{\mathbf{q}}}
\newcommand{\St}{\cev{\mathbf{S}}}

\newcommand{\wmsoc}{\mathbf{w}}
\newcommand{\wcsoc}{\mathbf{w}^{\text{re}}}
\newcommand{\wssoc}{\mathbf{w}^{\text{im}}}

\newcommand{\pd}{\text{p}^{\text{d}}}
\newcommand{\qd}{\text{q}^{\text{d}}}
\newcommand{\Sd}{\text{S}^{\text{d}}}
\newcommand{\pgmin}{\underline{\text{p}}^{\text{g}}}
\newcommand{\pgmax}{\overline{\text{p}}^{\text{g}}}
\newcommand{\qgmin}{\underline{\text{q}}^{\text{g}}}
\newcommand{\qgmax}{\overline{\text{q}}^{\text{g}}}
\newcommand{\vmmin}{\underline{\text{v}}}
\newcommand{\vmmax}{\overline{\text{v}}}
\newcommand{\dvamin}{\underline{\va}_{ij}}
\newcommand{\dvamax}{\overline{\va}_{ij}}

\newcommand{\gammaPwfr}{\vec{\gamma}^{\text{p,w}}}
\newcommand{\gammaQwfr}{\vec{\gamma}^{\text{q,w}}}
\newcommand{\gammaPwto}{\cev{\gamma}^{\text{p,w}}}
\newcommand{\gammaQwto}{\cev{\gamma}^{\text{q,w}}}
\newcommand{\gammaPrfr}{\vec{\gamma}^{\text{p,r}}}
\newcommand{\gammaQrfr}{\vec{\gamma}^{\text{q,r}}}
\newcommand{\gammaPrto}{\cev{\gamma}^{\text{p,r}}}
\newcommand{\gammaQrto}{\cev{\gamma}^{\text{q,r}}}
\newcommand{\gammaPifr}{\vec{\gamma}^{\text{p,i}}}
\newcommand{\gammaQifr}{\vec{\gamma}^{\text{q,i}}}
\newcommand{\gammaPito}{\cev{\gamma}^{\text{p,i}}}
\newcommand{\gammaQito}{\cev{\gamma}^{\text{q,i}}}

\newcommand{\lambdaP}{\lambda^{\text{p}}}
\newcommand{\lambdaQ}{\lambda^{\text{q}}}
\newcommand{\lambdaf}{\vec{\lambda}}
\newcommand{\lambdat}{\cev{\lambda}}
\newcommand{\lambdaPf}{\vec{\lambda}^{\,\text{p}}}
\newcommand{\lambdaPt}{\cev{\lambda}^{\,\text{p}}}
\newcommand{\lambdaQf}{\vec{\lambda}^{\,\text{q}}}
\newcommand{\lambdaQt}{\cev{\lambda}^{\,\text{q}}}

\renewcommand{\lambdaf}{\vec{\lambda}}
\renewcommand{\lambdat}{\cev{\lambda}}
\renewcommand{\lambdaPf}{\vec{\phantom{\lambda}}\hspace{-0.58em}{\lambda}^{\text{p}}}
\renewcommand{\lambdaPt}{\cev{\phantom{\lambda}}\hspace{-0.58em}{\lambda}^{\text{p}}}
\renewcommand{\lambdaQf}{\vec{\phantom{\lambda}}\hspace{-0.58em}{\lambda}^{\text{q}}}
\renewcommand{\lambdaQt}{\cev{\phantom{\lambda}}\hspace{-0.58em}{\lambda}^{\text{q}}}
\newcommand{\nuThermalfr}{\vec{\nu}}
\newcommand{\nuThermalSfr}{\vec{\nu}^{\, \text{s}}}
\newcommand{\nuThermalPfr}{\vec{\nu}^{\, \text{p}}}
\newcommand{\nuThermalQfr}{\vec{\nu}^{\, \text{q}}}
\newcommand{\nuThermalto}{\cev{\nu}}
\newcommand{\nuThermalSto}{\cev{\nu}^{\, \text{s}}}
\newcommand{\nuThermalPto}{\cev{\nu}^{\, \text{p}}}
\newcommand{\nuThermalQto}{\cev{\nu}^{\, \text{q}}}

\renewcommand{\nuThermalto}{\cev{\phantom{\nu}}\hspace{-0.56em}{\nu}}
\renewcommand{\nuThermalSto}{\cev{\phantom{\nu}}\hspace{-0.56em}{\nu}^{\,\text{s}}}
\renewcommand{\nuThermalPto}{\cev{\phantom{\nu}}\hspace{-0.56em}{\nu}^{\,\text{p}}}
\renewcommand{\nuThermalQto}{\cev{\phantom{\nu}}\hspace{-0.56em}{\nu}^{\,\text{q}}}
\newcommand{\omegaf}{\omega^{\text{f}}}
\newcommand{\omegat}{\omega^{\text{t}}}
\newcommand{\omegar}{\omega^{\text{re}}}
\newcommand{\omegai}{\omega^{\text{im}}}
\newcommand{\muPg}{{\mu}^{\text{pg}}}
\newcommand{\muPgMin}{\ubar{\mu}^{\text{pg}}}
\newcommand{\muPgMax}{\bar{\mu}^{\text{pg}}}
\newcommand{\muQg}{{\mu}^{\text{qg}}}
\newcommand{\muQgMin}{\ubar{\mu}^{\text{qg}}}
\newcommand{\muQgMax}{\bar{\mu}^{\text{qg}}}
\newcommand{\muWm}{{\mu}^{\text{w}}}
\newcommand{\muWmMin}{\ubar{\mu}^{\text{w}}}
\newcommand{\muWmMax}{\bar{\mu}^{\text{w}}}
\newcommand{\muAngleDiff}{{\mu}^{\theta}}
\newcommand{\muAngleDiffMin}{\ubar{\mu}^{\theta}}
\newcommand{\muAngleDiffMax}{\bar{\mu}^{\theta}}

\newcommand{\pdRef}{\bar{\text{p}}^{\text{d}}}
\newcommand{\qdRef}{\bar{\text{q}}^{\text{d}}}

\makeatletter
\let\old@ps@headings\ps@headings
\let\old@ps@IEEEtitlepagestyle\ps@IEEEtitlepagestyle
\def\psccfooter#1{%
    \def\ps@headings{%
        \old@ps@headings%
        \def\@oddfoot{\strut\hfill#1\hfill\strut}%
        \def\@evenfoot{\strut\hfill#1\hfill\strut}%
    }%
    \def\ps@IEEEtitlepagestyle{%
        \old@ps@IEEEtitlepagestyle%
        \def\@oddfoot{\strut\hfill#1\hfill\strut}%
        \def\@evenfoot{\strut\hfill#1\hfill\strut}%
    }%
    \ps@headings%
}
\makeatother

\begin{document}
\title{Dual Conic Proxies for AC Optimal Power Flow}

%% To specify the authors when (number of affiliations <= 2)
\author{
\IEEEauthorblockN{Guancheng Qiu, Mathieu Tanneau, Pascal Van Hentenryck}
\IEEEauthorblockA{H. Milton Stewart School of Industrial \& Systems Engineering, 
Georgia Institute of Technology, Atlanta, USA\\
qgc@gatech.edu, \{mathieu.tanneau, pascal.vanhentenryck\}@isye.gatech.edu}
}

\maketitle

\begin{abstract}
    In recent years, there has been significant interest in the development of machine learning-based optimization proxies for AC Optimal Power Flow (AC-OPF).
    Although significant progress has been achieved in predicting high-quality primal solutions, no existing learning-based approach can provide valid dual bounds for AC-OPF.   
    This paper addresses this gap by training optimization proxies for a convex relaxation of AC-OPF.
    Namely, the paper considers a second-order cone (SOC) relaxation of AC-OPF, and proposes \revision{a novel architecture} that embeds a fast, differentiable (dual) feasibility recovery, thus providing valid dual bounds.
    The paper combines this new architecture with a self-supervised learning scheme, which alleviates the need for costly training data generation.
    Extensive numerical experiments on medium- and large-scale power grids demonstrate the efficiency and scalability of the proposed methodology.
\end{abstract}


% Use this to place sponsorships
\thanksto{%
    \noindent This research is partly funded by NSF award 2112533 and ARPA-E PERFORM award AR0001136.
    \revision{The authors would like to thank the anonymous reviewers, whose valuable feedback helped improve this paper.}
}

\section{Introduction}
\label{sec:introduction}

    The AC Optimal Power Flow (AC-OPF) problem determines the most economical generation dispatch to meet current demand while satisfying physical and engineering constraints.
    Despite being fundamental to power systems operations, AC-OPF is seldom used in practice due to its nonlinear and non-convex nature, which makes it challenging to solve. Instead, most practitioners and market operators rely on the DC-OPF approximation. The computational limitations of AC-OPF, combined with the fact that similar instances need to be solved repeatedly, has spurred significant interest in developing optimization proxies for AC-OPF, i.e., machine learning (ML) models that approximate the input-output mapping of an AC-OPF solver. Once trained, AC-OPF proxies can produce high-quality solutions in milliseconds.

    Virtually all existing works on optimization proxies for DC- and AC-OPF focus on predicting primal solutions, i.e., generation dispatches, voltage setpoints, and power flows.
    Several approaches seek to improve the feasibility of predicted solutions by embedding constraints into the architecture and/or loss function \cite{fioretto2020predicting, donti2021dc3, park2023self, Chen2023_E2ELR}.
    Other approaches use feasibility restoration techniques, either as a post-processing step \cite{zamzam2020learning, Huang2022_DeepOPF-V} or as a differentiable restoration layer incorporated in an end-to-end manner \cite{donti2021dc3, Chen2023_E2ELR}.
    Restoring feasibility of a close-to-feasible solution is typically significantly faster than solving the original problem.

    Despite the impressive progress in developing optimization proxies for AC-OPF, a major limitation remains: \emph{existing optimization proxies do not provide certificates of (near-)optimality.}
    This contrasts with global optimization solvers, which provide both a primal solution and a certificate of (near-)optimality, i.e., a valid dual bound. The absence of optimality guarantees prevents the deployment of optimization proxies in real systems, because sub-optimal solutions may result in significant economic shortfalls, and electricity markets typically require (near-)optimal solutions \cite{MISO_BPM_002}.
    Recent works on neural network verification \cite{Nellikkath2021_WorstCaseViolations, Venzke2020_WorstCaseGuaranteesDNN} do offer worst-case performance guarantees for trained optimization proxies.
    Nevertheless, this approach is restricted to specific classes of models, does not scale well to large systems, and only provides a worst-case guarantee which may not be informative in practice.

    The paper addresses this fundamental gap by learning \emph{dual} optimization proxies that provide valid dual bounds in milliseconds.
    This is the first learning-based approach to provide valid dual bounds for AC-OPF, which is valuable in its own right.
    The paper makes the following contributions.
    \begin{enumerate}
        \item It proposes dual optimization proxies that leverage convex relaxations of AC-OPF to obtain \emph{valid} dual bounds in milliseconds.
        \item It develops a novel dual-feasible architecture, {\em Dual Conic Proxies} (\dcp{}), that uses conic duality to guarantee dual feasibility.
        \item It employs a self-supervised training algorithm to train \dcp{}, thus eliminating the need for costly, and potentially numerically unstable, optimization solvers.
        \item It conducts numerical experiments on large-scale power systems, which demonstrate that the \dcp{} approach is scalable and provides high-quality certified dual bounds compared to a conic optimization solver.
    \end{enumerate}

    The rest of the paper is organized as follows.
    Section \ref{sec:literature} surveys the relevant literature.
    Section \ref{sec:formulation} presents the AC-OPF problem, its second order cone (SOC) relaxation, and the dual of the SOC relaxation.
    Section \ref{sec:architecture} describes the dual-feasible architecture and self-supervised training.
    Section \ref{sec:results} reports numerical results, and Section \ref{sec:conclusion} concludes the paper.

\section{Related Work}
\label{sec:literature}

\subsection{Convex Relaxations of AC-OPF}
\label{sec:literature:convex}

    Convex relaxations of AC-OPF have attracted significant interest in the literature \cite{low2014convex1,low2014convex2}.
    They include, among others, second-order cone (SOC) \cite{Jabr2006_SOCRelaxationOPF}, semidefinite programming (SDP) \cite{Bai2008_SDPRelaxationOPF}, quadratic convex (QC) \cite{Coffrin2015_QCrelaxationOPF}, and moment-based relaxations \cite{Molzahn2014_Moment_OPF}.
    While the AC-OPF problem is non-convex and NP-hard \cite{lehmann2015ac}, convex relaxations can be solved efficiently, i.e., in polynomial time, thus providing certificates of (near-)optimality for AC-OPF solutions \cite{kocuk2016strong,Gopinath2020_ProvingGlovalACOPF}.

    While convex relaxations have received a lot of attention from the optimization and power systems communities, they have seldom been considered in ML settings.
    Infeasibility certificates from convex relaxations are used in \cite{Thams2020_EfficientDatabaseCreation,Venzke2021_EfficientCreationDataset} to accelerate the generation of datasets of AC-OPF instances.
    A classifier model is trained in \cite{Cengil2022_AccelerateGlobalOptimalSolutions} to predict which variables should be considered in an optimality-based bound tightening (OBBT) scheme, which results in improved computational performance.
    To the best of the authors' knowledge, the use of ML tools to predict solutions of convex relaxations of AC-OPF has not been considered before.
    {\em This paper is the first to propose ML-based dual conic proxies that are guaranteed to output dual-feasible solutions, thus providing valid lower bounds on the optimal value of AC-OPF.}

\subsection{Optimization Proxies for AC-OPF}
\label{sec:literature:proxies}

    The development of optimization proxies for OPF problems has seen a surge of interest in recent years, mostly centered around DC-OPF \cite{Nellikkath2021_WorstCaseViolations,deka2019learning,pan2020deepopf,chen2022learning,kim2022projection,li2023learning,Chen2023_E2ELR} and AC-OPF \cite{fioretto2020predicting, zamzam2020learning,chatzos2021spatial,pan2022deepopf,Mak2023_LearningACOPF_ADMM,park2023compact}.
    An overview of these efforts is presented next; an exhaustive review of optimization proxies for OPF problems is beyond the scope of this paper.
    
    Virtually all existing approaches on optimization proxies for OPF problems focus on predicting primal solutions, namely, generation dispatches, voltage setpoints, and power flows. Existing approaches have been applied to generating solutions for DC-OPF~\cite{deka2019learning,pan2020deepopf,Nellikkath2021_WorstCaseViolations,chen2022learning} and AC-OPF~\cite{fioretto2020predicting,zamzam2020learning,chatzos2021spatial,pan2022deepopf,park2023compact}.
    In particular, several techniques have been proposed to improve the feasibility of predicted solutions.
    Physics-informed architectures penalize constraint violations in the loss function \cite{fioretto2020predicting,donti2021dc3,chatzos2021spatial,Nellikkath2022_PINNACOPF,pan2022deepopf,park2023compact}.
    Equality completion is used in \cite{zamzam2020learning,pan2022deepopf,donti2021dc3} to ensure the predicted solution satisfy power flow equations; this step requires solving a system of linear or nonlinear equations.
    A post-processing feasibility restoration step is also employed in \cite{pan2020deepopf,Huang2022_DeepOPF-V}.
    Finally, a few papers \cite{Nellikkath2022_PINNACOPF,Mak2023_LearningACOPF_ADMM,park2023self,park2023compact} jointly predict primal and dual solutions.
    \cite{Nellikkath2021_WorstCaseViolations,Nellikkath2022_PINNACOPF} penalize the violation of KKT conditions during training, \cite{park2023self} consider an Augmented Lagrangian-based training scheme, and \cite{Mak2023_LearningACOPF_ADMM,park2023compact} feeds the predicted primal-dual solution as warm-start to an AC-OPF solver.
    With the exception of \cite{kim2022projection,li2023learning,Chen2023_E2ELR} which ensure feasibility of output DC-OPF solutions, the above approaches do not offer feasibility guarantees, unless a post-processing restoration step is employed.

    Traditionally, optimization proxies are trained using supervised learning, wherein one first generates a dataset of OPF instances and their solutions, then trains a model to minimize the distance between predicted and ground truth solutions.
    More recently, self-supervised approaches \cite{huang2021deepopf,owerko2022unsupervised,donti2021dc3,park2023self,Chen2023_E2ELR} directly minimize the objective value of the predicted solution, thereby eliminating the need for costly data generation.
    As pointed out in \cite{Chen2023_E2ELR}, a key challenge in self-supervised approaches is to ensure feasibility to avoid predicting trivial infeasible solutions.

\subsection{Optimization Proxy Verification}
\label{sec:literature:verification}
    
    Due to the risks involved in power system operations, neural network verification has been used to provide guarantees on output quality of OPF optimization proxies that are based on deep neural networks (DNN).   
    Mixed-integer-linear-programming (MILP)-based verification is used in  \cite{Venzke2020_WorstCaseGuaranteesDNN, Nellikkath2021_WorstCaseViolations} to provide guarantees of worst-case violation and worst-case sub-optimality of a DC-OPF proxy.
    The sub-optimality verification requires solving a bi-level problem, which is computationally expensive for large systems.
    \cite{Nellikkath2022_PINNACOPF} uses a mixed integer quadratic program (MIQP) to verify the worst-case constraint violation of a physics-informed neural network for AC-OPF.
    The work does not consider verification of optimality.
    \cite{Chevalier2023_GlobalPerformanceGuaranteesACOPF} develops a more tractable MIQP-based verification to verify the worst-case constraint violation for AC-OPF proxies, but does not consider the quality of the proposed approximation.
    \cite{Chevalier2023_GPUVerificationPowerSystems} accelerates the verification process for constraint violations (but not verification of optimality) by adapting it to a specialized, parallelized verification solver.
    The above methods are based on an approach which relies on an MILP encoding of the neural networks, so they only support models that are MILP-representable.

\section{AC-OPF and a Second-Order Cone Relaxation}
\label{sec:formulation}

Throughout the paper, the imaginary unit is denoted by $\im$, i.e., $\im^{2} \, {=} \, -1$.
The complex conjugate of $z \, {\in} \, \mathbb{C}$ is $z^{\star}$.
Additional notations used by the paper are defined in Tables \ref{tab:nomenclature:sets} and \ref{tab:nomenclature:parameters}.
For ease of reading and without loss of generality, the presentation assumes that exactly one generator is connected to each bus, and that all costs are linear.

\begin{table}[!t]
    \centering
    \caption{The Nomenclature of Sets.}
    \label{tab:nomenclature:sets}
    \resizebox{\columnwidth}{!}{
    \begin{tabular}{cc}
        \toprule
        Symbol & Description \\
        \midrule
        $\mathcal{N}$ & the set of buses, each represented as a node $i$ \\
        $\mathcal{E}$ & the set of oriented branches, each represented as a directed arc $(i, j)$ \\
        $\mathcal{E}^{+}_{i}$ & the set of arcs leaving bus $i$ \\
        $\mathcal{E}^{-}_{i}$ & the set of arcs entering bus $i$ \\
        \midrule
        $\mathcal{Q}^{n}$ & $\{x \in \mathbb{R}^{n} \mid x_{1} \geq \sqrt{x_{2}^{2} + ... + x_{n}^{2}}\}$ \\
        $\mathcal{Q}^{n}_{r}$ & $\{x \in \mathbb{R}^{n} \mid 2x_{1}x_{2} \geq x_{3}^{2} + ... + x_{n}^{2},\, x_{1}, x_{2} \geq 0\}$ \\
        \bottomrule
    \end{tabular}
    }
\end{table}

\begin{table}[!t]
    \centering
    \caption{The Nomenclature of Variables and Parameters in Formulations.}
    \label{tab:nomenclature:parameters}
    \resizebox{\columnwidth}{!}{
    \begin{tabular}{ll}
        \toprule
        Symbol & Description \\
        \midrule
        $\Sg_{i} = \pg_{i} + \im \qg_{i}$ & power generation at bus $i$ \\
        \midrule
        $\V_i = \vm_{i} \angle \va_{i}$ & voltage at bus $i$ \\
        \midrule
        $\Sf_{ij} = \pf_{ij} + \im \qf_{ij}$ & power flow from bus $i$ to bus $j$ on branch $(i,j)$ \\
        $\St_{ij} = \pt_{ij} + \im \qt_{ij}$ & power flow from bus $j$ to bus $i$ on branch $(i,j)$ \\
        \midrule
        $c_{i}$ & linear cost coefficient of power generation at bus $i$ \\
        \midrule
        $\Sd_{i} = \pd_{i} + \im \qd_{i}$ & power demand at bus $i$ \\
        \midrule
        $Y_{ij}$ & complex line admittance of branch $(i, j)$ \\
        $Y^{c}_{ij}$ & complex shunt admittance of branch $(i, j)$ \\
        \midrule
        $Y^{s}_{i}$ & complex admittance of shunt at bus $i$ \\
        \midrule
        $T_{ij}$ & complex transformer tap ratio of branch $(i, j)$ \\
        \midrule
        $\bar{s}_{ij}$ & thermal limit of branch $(i, j)$ \\
        $\pgmin_{i}$, $\pgmax_{i}$ & bounds on active power generation at bus $i$ \\
        $\qgmin_{i}$, $\qgmax_{i}$ & bounds on reactive power generation at bus $i$ \\
        $\vmmin_{i}$, $\vmmax_{i}$ & bounds on voltage magnitude at bus $i$ \\
        $\dvamin$, $\dvamax$ & bounds on angle difference of bus $i$ and bus $j$ \\
        \bottomrule
    \end{tabular}
    }
\end{table}

\subsection{The AC Optimal Power Flow Formulation}
\label{sec:formulation:AC}

The formulation of AC-OPF  considered in this paper is detailed in Model~\ref{model:ac_opf}.
The objective \eqref{eq:ACOPF:objective} minimizes total generation costs.
Constraint \eqref{eq:ACOPF:kirchhoff} enforces power balance (Kirchhoff's node law) at each bus.
Constraints \eqref{eq:ACOPF:ohm_fr}--\eqref{eq:ACOPF:ohm_to} enforce Ohm's law for forward and reverse power flows.
Constraints \eqref{eq:ACOPF:thermal_limits} enforce thermal limits on each branch.
Finally, constraints \eqref{eq:ACOPF:voltage_bounds}--\eqref{eq:ACOPF:reactive_dispatch_bounds} enforce minimum and maximum limits on nodal voltage magnitude, phase angle differences, and active/reactive generation.

\begin{model}[!t]
    \caption{The AC-OPF model}
	\label{model:ac_opf}
    \begin{subequations}
    \label{eq:ACOPF}
    \footnotesize
    \begin{align}
        \min \quad 
        & \sum_{i \in \mathcal{N}} c_{i} \pg_{i} \label{eq:ACOPF:objective}\\
        \textrm{s.t.} \quad
        & \Sg_{i} - \sum_{e \in \mathcal{E}^{+}_{i}} \Sf_{e} - \sum_{e \in \mathcal{E}^{-}_{i}} \St_{e} - (Y^{s}_{i})^{\star} |\V_{i}|^{2} = \Sd_{i}
            && \forall i \in \mathcal{N} \label{eq:ACOPF:kirchhoff} \\
        & \Sf_{ij} = (Y_{ij} + Y_{ij}^{c})^{\star} \frac{|\V_{i}|^{2}}{|T_{ij}|^{2}} - Y_{ij}^{\star}\frac{\V_{i} \V_{j}^{\star}}{T_{ij}} 
            && \forall ij \in \mathcal{E} \label{eq:ACOPF:ohm_fr}\\
        & \St_{ij} = (Y_{ij} + Y_{ji}^{c})^{\star} |\V_{j}|^{2} - Y_{ij}^{\star}\frac{\V_{i}^{\star} \V_{j}}{T_{ij}^{\star}} 
            && \forall ij \in \mathcal{E} \label{eq:ACOPF:ohm_to}\\
        & |\Sf_{ij}|, |\St_{ij}| \leq \bar{s}_{ij}
            && \forall ij \in \mathcal{E} \label{eq:ACOPF:thermal_limits} \\
        & \vmmin_{i} \leq |\V_{i}| \leq \vmmax_{i} 
            && \forall i \in \mathcal{N} \label{eq:ACOPF:voltage_bounds}\\
        & \dvamin \leq \va_{i} - \va_{j} \leq \dvamax 
            && \forall ij \in \mathcal{E} \label{eq:ACOPF:angle_diff_bounds}\\
        & \pgmin_{i} \leq \pg_{i} \leq \pgmax_{i}
            && \forall i \in \mathcal{N} \label{eq:ACOPF:active_dispatch_bounds}\\
        & \qgmin_{i} \leq \qg_{i} \leq \qgmax_{i} 
            && \forall i \in \mathcal{N} \label{eq:ACOPF:reactive_dispatch_bounds}
    \end{align}
    \end{subequations}
\end{model}

\subsection{The Second-Order Cone Relaxation of the AC-OPF}
\label{sec:formulation:SOC}

The second-order cone (SOC) relaxation of AC-OPF proposed by Jabr in \cite{Jabr2006_SOCRelaxationOPF} is obtained by introducing changes of variables
\begin{align}
    \wmsoc_{i}  &= \vm_{i}^{2} && \forall i \in \mathcal{N} \\
    \wcsoc_{ij} &= \vm_{i} \vm_{j} \cos(\va_{i} - \va_{j}) && \forall ij \in \mathcal{E} \\
    \wssoc_{ij} &= \vm_{i} \vm_{j} \sin(\va_{i} - \va_{j}) && \forall ij \in \mathcal{E}
\end{align}
together with the additional non-convex constraints
\begin{align}
\label{eq:quadratic-constraint}
    (\wcsoc_{ij})^{2} + (\wssoc_{ij})^{2} = \wmsoc_{i} \wmsoc_{j} \qquad \forall ij \in \mathcal{E}.
\end{align}
The non-convex quadratic constraints \eqref{eq:quadratic-constraint} are then relaxed as
\begin{align}
\label{eq:jabr_inequality}
    (\wcsoc_{ij})^{2} + (\wssoc_{ij})^{2} \leq \wmsoc_{i} \wmsoc_{j} \qquad \forall ij \in \mathcal{E}.
\end{align}

\begin{model}[!t]
\caption{The SOC-OPF Model.}
\label{model:SOC-OPF}
\begin{subequations}
\label{eq:SOCOPF}
\footnotesize
\begin{align}
    \min \quad 
        & \sum_{i \in \mathcal{N}} c_{i} \pg_{i} \\
    \textrm{s.t.} \quad 
    & 
        \pg_{i} - \pd_{i} - g_{i}^{s} \wmsoc_{i} = \sum_{e \in \mathcal{E}^{+}_{i}} \pf_{e} - \sum_{e \in \mathcal{E}^{-}_{i}} \pt_{e} 
        && \forall i {\, {\in} \,} \mathcal{N} 
        && [\lambdaP_{i}]
        \label{eq:SOCOPF:kirchhoff:active} \\
    & 
        \qg_{i} - \qd_{i} + b_{i}^{s} \wmsoc_{i} = \sum_{e \in \mathcal{E}^{+}_{i}} \qf_{e} - \sum_{e \in \mathcal{E}^{-}_{i}} \qt_{e} 
        && \forall i {\, {\in} \,} \mathcal{N} 
        && [\lambdaQ_{i}]
        \label{eq:SOCOPF:kirchhoff:reactive} \\
    &
        \pf_{ij} =
        \gammaPwfr_{ij} \wmsoc_{i} + \gammaPrfr_{ij} \wcsoc_{ij} + \gammaPifr_{ij} \wssoc_{ij}
        && \forall ij {\, {\in} \,} \mathcal{E}  
        && [\lambdaPf_{ij}]
        \label{eq:SOCOPF:ohm:active:fr} \\
    &
        \pt_{ij} =
        \gammaPwto_{ij} \wmsoc_{j} + \gammaPrto_{ij} \wcsoc_{ij} + \gammaPito_{ij} \wssoc_{ij}
        && \forall ij {\, {\in} \,} \mathcal{E} 
        && [\lambdaPt_{ij}]
        \label{eq:SOCOPF:ohm:active:to} \\
    &
        \qf_{ij} =
        \gammaQwfr_{ij} \wmsoc_{i} + \gammaQrfr_{ij} \wcsoc_{ij} + \gammaQifr_{ij} \wssoc_{ij}
        && \forall ij {\, {\in} \,} \mathcal{E} 
        && [\lambdaQf_{ij}]
        \label{eq:SOCOPF:ohm:reactive:fr} \\
    &
        \qt_{ij} =
        \gammaQwto_{ij} \wmsoc_{j} + \gammaQrto_{ij} \wcsoc_{ij} + \gammaQito_{ij} \wssoc_{ij}
        && \forall ij {\, {\in} \,} \mathcal{E} 
        && [\lambdaQt_{ij}]
        \label{eq:SOCOPF:ohm:reactive:to}  \\
    & 
        (\bar{s}_{ij}, \pf_{ij}, \qf_{ij}) \in \mathcal{Q}^{3}
        && \forall ij {\, {\in} \,} \mathcal{E} 
        && [\nuThermalfr_{ij}]
        \label{eq:SOCOPF:thermal:fr} \\
    & 
        (\bar{s}_{ij}, \pt_{ij}, \qt_{ij}) \in \mathcal{Q}^{3}
        && \forall ij {\, {\in} \,} \mathcal{E} 
        && [\nuThermalto_{ij}]
        \label{eq:SOCOPF:thermal:to} \\
    & 
        \tan(\dvamin) \wcsoc_{ij} \leq \wssoc_{ij} \leq \tan(\dvamax)\wcsoc_{ij} 
        && \forall ij {\, {\in} \,} \mathcal{E} 
        && [\muAngleDiff_{ij}]
        \label{eq:SOCOPF:angle_diff}\\
    & 
        (\frac{\wmsoc_{i}}{\sqrt{2}}, \frac{\wmsoc_{j}}{\sqrt{2}}, \wcsoc_{ij}, \wssoc_{ij}) \in \mathcal{Q}_{r}^{4} 
        && \forall ij {\, {\in} \,} \mathcal{E}  
        && [\omega_{ij}]
        \label{eq:SOCOPF:jabr_soc} \\
    & 
        \vmmin_{i}^{2} \leq \wmsoc_{i} \leq \vmmax_{i}^{2} 
        && \forall i {\, {\in} \,} \mathcal{N} 
        && [\muWm_{i}]
        \label{eq:SOCOPF:wm:bounds}\\
    & 
        \pgmin_{i} \leq \pg_{i} \leq \pgmax_{i} 
        && \forall i {\, {\in} \,} \mathcal{N} 
        && [\muPg_{i}]
        \label{eq:SOCOPF:pg:bounds}\\
    & 
        \qgmin_{i} \leq \qg_{i} \leq \qgmax_{i}
        && \forall i {\, {\in} \,} \mathcal{N}  
        && [\muQg_{i}]
        \label{eq:SOCOPF:qg:bounds}
    \end{align}
\end{subequations}
\end{model}

The resulting SOC-OPF formulation (in real notations) is presented in Model \ref{model:SOC-OPF}, with each constraint's dual variable indicated in brackets.
Constraints \eqref{eq:SOCOPF:kirchhoff:active}--\eqref{eq:SOCOPF:kirchhoff:reactive} enforce active and reactive power balance at each bus.
Constraints \eqref{eq:SOCOPF:ohm:active:fr}--\eqref{eq:SOCOPF:ohm:reactive:to} formulate Ohm's law for active/reactive forward and reverse power flows, where constants $\vec{\gamma}, \cev{\gamma}$ are obtained from Eq. \eqref{eq:ACOPF:ohm_fr}--\eqref{eq:ACOPF:ohm_to}.
Conic constraints \eqref{eq:SOCOPF:thermal:fr}--\eqref{eq:SOCOPF:thermal:to} enforce thermal limits on each branch's forward and reverse power flow.
Constraints \eqref{eq:SOCOPF:angle_diff} ensure phase angle difference limits on each branch, and constraint \eqref{eq:SOCOPF:jabr_soc} is Jabr's inequality \eqref{eq:jabr_inequality} in conic form.
Finally, constraints \eqref{eq:SOCOPF:wm:bounds}--\eqref{eq:SOCOPF:qg:bounds} enforce minimum and maximum limits on nodal voltage magnitude and active/reactive dispatch.

\begin{model*}[!t]
\caption{The DSOC-OPF Model.}
\label{model:DSOC-OPF}
\begin{subequations}
    \footnotesize
    \begin{align}
        \max_{\lambda, \mu, \nu} \quad 
        & \sum_{i \in \mathcal{N}} \left(
            \pd_{i} \lambdaP_{i}
            + \qd_{i} \lambdaQ_{i}
            + \pgmin_{i} \muPgMin_{i} - \pgmax_{i} \muPgMax_{i}
            + \qgmin_{i} \muQgMin_{i} - \qgmax_{i} \muQgMax_{i}
            + \vmmin_{i}^{2} \muWmMin_{i} - \vmmax_{i}^{2} \muWmMax_{i}
        \right)
        - \sum_{e \in \mathcal{E}} \bar{s}_{e} \left( \nuThermalSfr_{e} + \nuThermalSto_{e} \right)
        \label{eq:DSOC:obj}\\
        s.t. \quad
        & 
            \lambdaP_{i} + \muPgMin_{i} - \muPgMax_{i} = c_{i}
            && \forall i \in \mathcal{N}
            \label{eq:DSOC:pg}\\
        & 
            \lambdaQ_{i} + \muQgMin_{i} - \muQgMax_{i} = 0
            && \forall i \in \mathcal{N}
            \label{eq:DSOC:qg}\\
        & 
            -\lambdaP_{i} - \lambdaPf_{ij} + \nuThermalPfr_{ij} = 0
            && \forall ij \in \mathcal{E}
            \label{eq:DSOC:pf}\\
        & 
            -\lambdaQ_{i} - \lambdaQf_{ij} + \nuThermalQfr_{ij} = 0
            && \forall ij \in \mathcal{E}
            \label{eq:DSOC:qf}\\
        & 
            -\lambdaP_{j} - \lambdaPt_{ij }+ \nuThermalPto_{ij} = 0
            && \forall ij \in \mathcal{E}
            \label{eq:DSOC:pt}\\
        & 
            -\lambdaQ_{j} - \lambdaQt_{ij} + \nuThermalQto_{ij} = 0
            && \forall ij \in \mathcal{E}
            \label{eq:DSOC:qt}\\
        &
            -g_{i}^{s} \lambdaP_{i} + b_{i}^{s} \lambdaQ_{i}
            + \sum_{e \in \mathcal{E}^{+}_{i}} \left(
                \gammaPwfr_{e} \lambdaPf_{e} 
                + \gammaQwfr_{e} \lambdaQf_{e} 
                + \frac{\omegaf_{e}}{\sqrt{2}}
            \right)
            + \sum_{e \in \mathcal{E}^{-}_{i}} \left(
                \gammaPwto_{e} \lambdaPt_{e} 
                + \gammaQwto_{e} \lambdaQt_{e}
                + \frac{\omegat_{e}}{\sqrt{2}}
            \right)
            + \muWmMin_{i} - \muWmMax_{i}
            = 0
            && \forall i \in \mathcal{N}
            \label{eq:DSOC:wm}\\
        &
            \gammaPrfr_{ij} \lambdaPf_{ij}
            + \gammaPrto_{ij} \lambdaPt_{ij}
            + \gammaQrfr_{ij} \lambdaQf_{ij}
            + \gammaQrto_{ij} \lambdaQt_{ij}
            - \tan(\dvamin) \muAngleDiffMin_{ij}
            + \tan(\dvamax) \muAngleDiffMax_{ij}
            + \omegar_{ij}
            = 0
            && \forall ij \in \mathcal{E}
            \label{eq:DSOC:wr}\\
        &
            \gammaPifr_{ij} \lambdaPf_{ij}
            + \gammaPito_{ij} \lambdaPt_{ij}
            + \gammaQifr_{ij} \lambdaQf_{ij}
            + \gammaQito_{ij} \lambdaQt_{ij}
            + \muAngleDiffMin_{ij}
            - \muAngleDiffMax_{ij}
            + \omegai_{ij}
            = 0
            && \forall ij \in \mathcal{E}
            \label{eq:DSOC:wi}\\
        % Domain of dual variables
        & 
            \muPgMin, \muPgMax, \muQgMin, \muQgMax, \muWmMin, \muWmMax, \muAngleDiffMin, \muAngleDiffMax \geq 0
            && 
            \label{eq:DSOC:non_negative}\\
        & 
            \nuThermalfr_{ij} = (\nuThermalSfr_{ij}, \nuThermalPfr_{ij}, \nuThermalQfr_{ij}) \in \mathcal{Q}^{3},
            \nuThermalto_{ij}  = (\nuThermalSto_{ij}, \nuThermalPto_{ij}, \nuThermalQto_{ij})\in \mathcal{Q}^{3}
            && \forall ij \in \mathcal{E}
            \label{eq:DSOC:cone:nu}\\
        & 
            \omega_{ij} = \left(\omegaf_{ij}, \omegat_{ij}, \omegar_{ij}, \omegai_{ij} \right) \in \mathcal{Q}_{r}^{4},
            && \forall ij \in \mathcal{E}
            \label{eq:DSOC:cone:omega}
    \end{align}
\end{subequations}
\end{model*}

This paper focuses on the dual problem DSOC-OPF, formulated in Model \ref{model:DSOC-OPF}.
By convention, dual variables corresponding to two-sided linear inequality constraints are denoted by $\ubar{\mu}$ and $\bar{\mu}$, respectively.
For instance, $\muPgMin, \muPgMax$ denote the dual variables associated to active power generation lower and upper bounds.
Also note that the conic dual variables $\nuThermalfr, \nuThermalto, \omega$ associated to constraints \eqref{eq:SOCOPF:thermal:fr}, \eqref{eq:SOCOPF:thermal:to}, \eqref{eq:SOCOPF:jabr_soc} are vectors, as stated in constraints \eqref{eq:DSOC:cone:nu}, \eqref{eq:DSOC:cone:omega}.
It is important to note that the DSOC-OPF is a second-order conic optimization problem and that, by weak duality, any feasible solution for the DSOC-OPF yields a valid lower bound on the objective value of the SOC-OPF and, in turn, of the AC-OPF.
Dual problems, such as the DSOC-OPF, have received very little attention in the literature; to the best of the author's knowledge, {\em this paper is the first to apply machine learning techniques to the DSOC-OPF}.

\section{The Dual-Feasible Proxy Architecture}
\label{sec:architecture}

\begin{figure}[!t]
    \centering
    \includegraphics[width=\columnwidth]{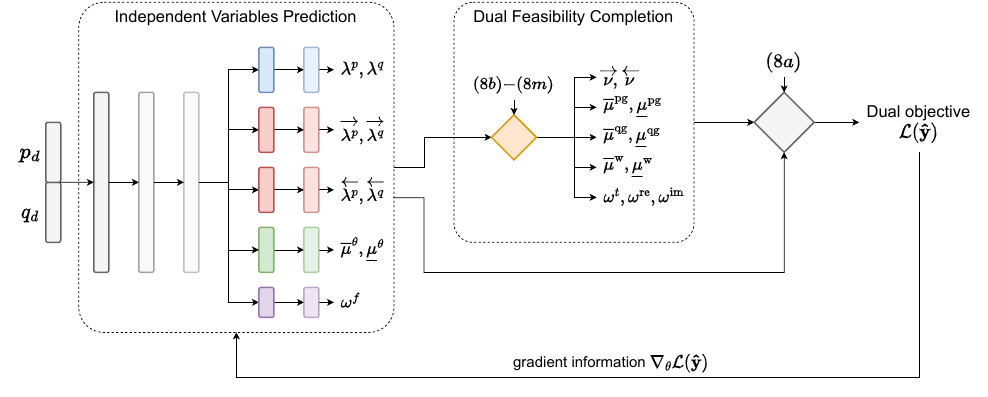}
    \caption{The Proposed Dual Conic Proxy Architecture.}
    \label{fig:DCP}
\end{figure}

This section presents the proposed DCP architecture, illustrated
in Figure \ref{fig:DCP}. DCP predicts a subset of variables (the
\emph{independent} variables) before applying a completion step that
computes values of the remaining variables (the \emph{dependent}
variables) that yields a dual-feasible solution.  At a high level
(and with a slight approximation), the prediction step of DCP delivers
dual values for physical and phase angle constraints, while its
completion step produces values for the remaining duals. {\em The
  completion step of DCP is a key contribution of the paper. Contrary
  to existing completion approaches, e.g., \cite{donti2021dc3}, that
  only use equality constraints, DCP leverages fundamental properties
  of dual-optimal solutions.}
  
In the following, $\mathbf{x}$ (resp. $\mathbf{y}$) denotes the vector of all
primal (resp. dual) variables.
It is easy to verify that SOC-OPF is always bounded, and that DSOC-OPF is always strictly feasible.
In addition, the paper assumes that SOC-OPF is strictly feasible.
Therefore, by strong conic duality (see Theorem 1.4.2 in \cite{ben2001lectures}), both problems are solvable, and any primal-dual optimal solution satisfies complementary slackness.

\subsection{Dual Feasibility Completion}
\label{sec:architecture:feasibility}

The dual-feasibility completion takes as inputs a prediction for a set
of \emph{independent} variables, and outputs
values for the remaining \emph{dependent} variables that ensure that
Constraints \eqref{eq:DSOC:pg}--\eqref{eq:DSOC:cone:omega} are
satisfied by the resulting assignment of all variables. Existing
completion approaches, e.g., \cite{donti2021dc3}, only use equality
constraints.  In contrast, DCP also leverages properties of
dual-optimal solutions, which are formalized in the following Lemmas.

\begin{lemma}
    \label{lemma:dual_optimal_slacks}
    Let $\mathbf{y}$ be dual-optimal.
    Then,
    \begin{align*}
        \forall i \in \mathcal{N}, \ 
        \muPgMin_{i} \times \muPgMax_{i} =
        \muQgMin_{i} \times \muQgMax_{i} = 
        \muWmMin_{i} \times \muWmMax_{i} = 0.
    \end{align*}
\end{lemma}
\begin{proof}
The proof is given for $\muPgMin, \muPgMax$; it is similar for
$\muQgMin, \muQgMax$ and $\muWmMin, \muWmMax$.  Let $i \, {\in} \,
\mathcal{N}$, and assume $\muPgMin_{i} \, {\times} \, \muPgMax_{i} \,
        {>} \, 0$, i.e., $\muPgMin_{i}, \muPgMax_{i} \, {>} \,0$
        because of \eqref{eq:DSOC:non_negative}.  Define $\delta
        = \min \{\muPgMin_{i}, \muPgMax_{i}\} > 0$.  Then,
        $(\muPgMin_{i} {-} \delta, \muPgMax_{i} {-} \delta)$ yields a
        dual-feasible solution, and increases the objective by
        $(\pgmax_{i} {-} \pgmin_{i}) \delta \, {>} \, 0$, which
        contradicts the assumption that $\mathbf{y}$ is optimal.
\end{proof}

\begin{lemma}
    \label{lemma:nu_thermal_on_face}
    Let $\mathbf{y}$ be dual-optimal. Then,
    \begin{align*}
        \forall ij \in \mathcal{E}, \ 
        \nuThermalSfr_{ij} = \sqrt{(\nuThermalPfr_{ij})^2 + (\nuThermalQfr_{ij})^2},
        \ 
        \nuThermalSto_{ij} = \sqrt{(\nuThermalPto_{ij})^2 + (\nuThermalQto_{ij})^2}.
    \end{align*}
\end{lemma}
\begin{proof}
    Variables $\nuThermalSfr_{ij}$, $\nuThermalSto_{ij}$ only appear in the SOC constraints \eqref{eq:DSOC:cone:nu}, and have negative objective coefficients.
\end{proof}

\begin{lemma}
    \label{lemma:omega_on_face}
    Let $\mathbf{y}$ be dual-optimal. Then,
    \begin{align}
        \label{eq:omega_on_face}
        \forall ij \in \mathcal{E}, \ 2 \omegaf_{i} \omegat_{j} = (\omegar_{ij})^{2} + (\omegai_{ij})^{2}
    \end{align}
\end{lemma}
\begin{proof}
    The result is a direct consequence of conic complementary slackness which, in the current setting, reads
    \begin{align}
        \label{eq:complementary_slackness:omega}
         (\frac{\wmsoc_{i}}{\sqrt{2}}, \frac{\wmsoc_{j}}{\sqrt{2}}, \wcsoc_{ij}, \wssoc_{ij})^{\top}
        \left(\omegaf_{ij}, \omegat_{ij}, \omegar_{ij}, \omegai_{ij} \right) = 0
    \end{align}
    In practice, $\vmmin \, {>} \, 0$, so constraints \eqref{eq:SOCOPF:wm:bounds} impose $\wmsoc_{i}, \wmsoc_{j} \, {>} \, 0$, which yields $(\frac{\wmsoc_{i}}{\sqrt{2}}, \frac{\wmsoc_{j}}{\sqrt{2}}, \wcsoc_{ij}, \wssoc_{ij}) \,{\neq} \, 0$.
    So for complementary slackness \eqref{eq:complementary_slackness:omega} to hold, $\left(\omegaf_{ij}, \omegat_{ij}, \omegar_{ij}, \omegai_{ij} \right)$ cannot be strictly conic-feasible, i.e., $\left(\omegaf_{ij}, \omegat_{ij}, \omegar_{ij}, \omegai_{ij} \right) \, {\notin} \, \text{int}(\mathcal{Q}_{r}^{4})$.
    Thus, constraint \eqref{eq:DSOC:cone:omega} is active, i.e., $2\omegaf_{ij} \omegat_{ij} \, {=} \, (\omegar_{ij})^{2} {+}(\omegai_{ij})^{2}$.
\end{proof}

\begin{algorithm}[!t]
    \caption{The DCP Methodology for DSOC-OPF.}
    \label{algo:DCP}
    \SetAlgoLined
    Predict $\lambdaP, \lambdaQ$, $\lambdaPf, \lambdaQf, \lambdaPt, \lambdaQt$, $\muAngleDiffMin, \muAngleDiffMax$, $\omegaf$ \\
    Recover $\nuThermalPfr, \nuThermalQfr, \nuThermalPto, \nuThermalQto$ using constraints \eqref{eq:DSOC:pf}--\eqref{eq:DSOC:qt} \label{algo:recover_nu} \\
    Recover $\nuThermalSfr, \nuThermalSto$ using Lemma \ref{lemma:nu_thermal_on_face} \\    
    Recover $\omegar, \omegai$ using constraints \eqref{eq:DSOC:wr}--\eqref{eq:DSOC:wi} \\
    Recover $\omegat$ from $\omegaf, \omegar, \omegai$ using Lemma \ref{lemma:omega_on_face} \\
    Recover $\muPgMin, \muPgMax, \muQgMin, \muQgMax, \muWmMin, \muWmMax$ via \eqref{eq:DSOC:pg}, \eqref{eq:DSOC:qg}, \eqref{eq:DSOC:wm} \\
\end{algorithm}

The DCP framework applied to DSOC-OPF is presented in Algorithm
\ref{algo:DCP}. Lemma \ref{lemma:dual_optimal_slacks} can be used to
recover variables $\muPgMin, \muPgMax$, $\muQgMin, \muQgMax, \muWmMin,
\muWmMax$ using constraints \eqref{eq:DSOC:pg}, \eqref{eq:DSOC:qg},
and \eqref{eq:DSOC:wm}.  Indeed, it suffices to impose $\muPgMin_{i}
\, {=} \, \max \{0, c_{i} {-} \lambdaP_{i}\}$ and $\muPgMax_{i} \, {=}
\, \max\{0, \lambdaP_{i} {-} c_{i} \}$; variables $\muQgMin, \muQgMax,
\muWmMin, \muWmMax$ are recovered in a similar fashion.  Lemma
\ref{lemma:nu_thermal_on_face} can be used to recover $\nuThermalSfr,
\nuThermalSto$ from $\nuThermalPfr, \nuThermalQfr, \nuThermalPto,
\nuThermalQto$.  Lemma \ref{lemma:omega_on_face} further eliminates
$|\mathcal{E}|$ degrees of freedom by proving that
\eqref{eq:DSOC:cone:omega} is always active at a dual optimum.  This
makes it possible to first recover $\omegar, \omegai$ using constraints
\eqref{eq:DSOC:wr}--\eqref{eq:DSOC:wi}, then $\omegat$ from $\omegaf,
\omegar, \omegai$.

\revision{%
Denote by $\xi^{\text{ind}} \, {=} \, (\lambdaP, \lambdaQ, \lambdaPf, \lambdaQf, \lambdaPt, \lambdaQt, \muAngleDiffMin, \muAngleDiffMax, \omegaf)$ and by $\xi^{\text{dep}} \, {=} \, (\nuThermalfr, \nuThermalto, \omegat, \omegar, \omegai, \muPgMin, \muPgMax, \muWmMin, \muWmMax)$ the set of independent and dependent variables, respectively.
Recall that Algorithm \ref{algo:DCP} allows to recover dependent variables $\hat{\xi}^{\text{dep}}$ from the predicted independent variables $\hat{\xi}^{\text{ind}}$.
An important property of the recovered solution, is that it attains the best possible dual bound among all possible recovered solutions.
\begin{theorem}
    \label{thm:recovery_best_bound}
    Let $\hat{\xi}^{\text{ind}}$ be fixed, and define $\hat{\xi}^{\text{dep}}$ the recovered dependent variables from Algorithm \ref{algo:DCP}.
    Then
    \begin{align}
        \label{eq:recovery_lagrange_bound}
        \hat{\xi}^{\text{dep}} 
        \in 
        \arg \max_{\xi^{\text{dep}}} \left\{ 
            \eqref{eq:DSOC:obj}
        \ \middle| \ 
            \eqref{eq:DSOC:pg}-\eqref{eq:DSOC:cone:omega}, \ 
            \xi^{\text{ind}} = \hat{\xi}^{\text{ind}}
        \right\}
    \end{align}
\end{theorem}
\begin{proof}
    The proof follows the same steps as the proofs of Lemmas  \ref{lemma:dual_optimal_slacks}--\ref{lemma:omega_on_face} and outlined in Algorithm \ref{algo:DCP}.
    Namely, $\nuThermalPfr, \nuThermalQfr, \nuThermalPto, \nuThermalQto$ are uniquely obtained from \eqref{eq:DSOC:pf}--\eqref{eq:DSOC:qt}, followed by $\nuThermalSfr, \nuThermalSto$ using the {\em optimality conditions} of Lemma \ref{lemma:nu_thermal_on_face}.
    Then, $\omegar, \omegai$ are uniquely obtained from constraints \eqref{eq:DSOC:wr}--\eqref{eq:DSOC:wi}, followed by $\omegat$ using the optimality conditions of Lemma \ref{lemma:omega_on_face}, i.e., that the conic constraint \eqref{eq:DSOC:cone:omega} must be active in any optimal solution of Problem \eqref{eq:recovery_lagrange_bound}.
    Finally, $\muPgMin, \muPgMax, \muWmMin, \muWmMax$ are recovered using the optimality conditions of Lemma \ref{lemma:dual_optimal_slacks}.
    This concludes the proof by showing that the recovered dual solution is optimal for Problem \eqref{eq:recovery_lagrange_bound}.
\end{proof}}

Note that the above dual completion procedure is not unique.  Indeed,
DSOC-OPF contains $8|\mathcal{N}| \,{+}\, 16|\mathcal{E}|$ degrees of
freedom (i.e., scalar variables), $3|\mathcal{N}| \,{+}\, 6|\mathcal{E}|$
linear equality constraints, $6|\mathcal{N}| \,{+}\, 2|\mathcal{E}|$ linear
inequality constraints, and $3|\mathcal{E}|$ SOC constraints.  Each
equality constraint eliminates one degree of freedom.  The results of
Lemmas \ref{lemma:dual_optimal_slacks}--\ref{lemma:omega_on_face}
eliminate another $3 |\mathcal{N}|$, $2 |\mathcal{E}|$, and
$|\mathcal{E}|$ degrees of freedom, respectively.  Therefore, any
valid completion procedure should be given $2 |\mathcal{N}|
\, {+} \, 7 |\mathcal{E}|$ variables, and recover the remaining $6
|\mathcal{N}| \,{+}\, 9 |\mathcal{E}|$ via equality constraints and Lemmas
\ref{lemma:dual_optimal_slacks}--\ref{lemma:omega_on_face}. This
leaves ample freedom in the choice of dependent and independent
variables, which can significantly impact the performance of
the resulting dual conic proxy. For instance, an alternative approach,
explored in Section \ref{sec:results}, is to initially
predict $\nuThermalPfr$, $\nuThermalQfr$, $\nuThermalPto$, $\nuThermalQto$,
and recover $\lambdaPf$, $\lambdaQf$, $\lambdaPt$, $\lambdaQt$ at step \ref{algo:recover_nu} of Algorithm \ref{algo:DCP}.

\subsection{Rectangular vs Polar Representation}
\label{sec:architecture:rect_vs_polar}

Another modifiable design in a DSOC-OPF proxy is whether to
represent complex quantities in rectangular or polar form.  Namely,
one may represent $z \, {\in} \, \mathbb{C}$ as $z \, {=} \, p \, {+}
\, \im q$ where $p, q \, {\in} \, \mathbb{R}$, or as $z \, {=} \, \rho
\angle \theta$ where $\rho \geq 0$, $\theta \in \mathbb{R}$.  While
optimization formulations may require rectangular or polar form to
ensure convexity, there is no such restriction when designing ML
models, which are typically non-convex.  This work has found that
representing $(\omegaf,\omegat)$ in polar form substantially improves
optimality of the proxy (see Section \ref{sec:results}).  {\em To the
  best of the authors' knowledge, the impact of polar vs rectangular
  representations on the performance of OPF proxies has not been
  studied before.}

The dual feasibility completion outlined in Algorithm 
\ref{algo:DCP} recovers $\omegat$ from $\omegaf,
\omegar, \omegai$ using Eq. \eqref{eq:omega_on_face}.  However, this
approach is numerically unstable if $\omegaf \, {\ll} \, 1$, and
precludes $(\omegaf, \omegat) \, {=} \, (0, 0)$, which may hold in a
dual-optimal solution.  Therefore, the paper elects to predict both
$\omegaf, \omegat$, then recover $(\tilde{\omega}^{\text{f}},
\tilde{\omega}^{\text{t}}) \, {=} \, (\omegaf {+} \delta\omega,
\omegat {+} \delta \omega)$ that satisfies
Eq. \eqref{eq:omega_on_face}.  Thereby, $\delta \omega$ is obtained by
solving a second-order equation. This approach represents
$(\omegaf, \omegat)$ in rectangular form.

The result of Lemma \ref{lemma:omega_on_face} shows that $(\omegaf,
\omegat)$ reside on a hyperbola, which can be represented using polar
coordinates, thus suggesting the following change of variable.  Let
$(\omegaf, \omegat) \, {=} \, (\rho \cos \phi, \rho \sin \phi)$ where
$\rho \geq 0$ and $\phi \in [0, \nicefrac{\pi}{2}]$.  Substituting
into Eq. \eqref{eq:omega_on_face}, this yields
\begin{align}
    \rho = \sqrt{\frac{{(\omegar)^2 + (\omegai)^{2}}}{{\sin(2\phi)}}}.
\end{align}
Preliminary analysis of optimal solutions provided by optimization solver reveals that the distribution of $\phi$ angles,
across multiple instances, has a mean close to $\pi/4$ and very low
variance, which makes it easy to learn.  This approach also has the
advantage of predicting $|\mathcal{E}|$ fewer independent variables.

\subsection{The Dual Conic Proxy Architecture and Training}
\label{sec:architecture:ML}

The DCP architecture, illustrated in Figure \ref{fig:DCP}, is flexible
and can be easily adapted to different ML models, different sets of
independent variables, or different internal representations as
outlined in Sections \ref{sec:architecture:feasibility} and
\ref{sec:architecture:rect_vs_polar}.

The first component of the DCP model is a Deep Neural Network (DNN)
model that takes, as inputs, active and reactive loads $\pd, \qd$, and
outputs independent variables.  The DNN architecture considered in the
experiments consists of an initial set of fully-connected layers,
followed by sub-networks, each of which outputs a subset of the
independent variables.  The use of sub-networks makes the overall
architecture more scalable, accounting for the large number of
independent variables ($2 |\mathcal{N}| + 7 |\mathcal{E}|$) to
predict.

The second component of the DCP model is the dual feasibility
completion.  This module takes, as inputs, the independent
variables, as well as relevant problem data, and applies Algorithm
\ref{algo:DCP} to recover the values of all dependent variables.  The
completed solution $\hat{\mathbf{y}}$ satisfies dual feasibility by
design.  The dual feasibility completion uses closed-form formulae,
which are supported by most ML libraries, and are differentiable
almost everywhere.  The latter allows for computing gradient information
through back-propagation.

The DCP model is trained in a self-supervised way, using the objective
function of the optimization problem for the loss function
$\mathcal{L}(\hat{\mathbf{y}})$. The loss function is back-propagated
through the completion layer back to the weights of the DNN in
an end-to-end manner. Self-supervised training alleviates the need for
offline data generation, and has been shown to provide higher-quality
solutions compared to supervised training \cite{Chen2023_E2ELR}.  As
pointed out in \cite{Chen2023_E2ELR}, ensuring feasibility is key to
the performance of self-supervised training.

\section{Experimental Results}
\label{sec:results}

\newcommand{\ieeeXXS}{\texttt{ieee14}}
\newcommand{\ieeeXS}{\texttt{ieee118}}
\newcommand{\ieeeS}{\texttt{ieee300}}
\newcommand{\pegaseSmall}{\texttt{pegase1354}}
\newcommand{\pegaseMedium}{\texttt{pegase2869}}

\subsection{Data Generation}
\label{sec:results:data_generation}

The paper conducts experiments on several systems from the PGLib
library v21.07 \cite{babaeinejadsarookolaee2019power}.  Table
\ref{tab:results:pglib} reports, for each system, the number of buses
($|\mathcal{N}|$) and branches ($|\mathcal{E}|$), the number of independent variables
(\#indep, equals $2|\mathcal{N}|{+}7|\mathcal{E}|$), the total active power demand
($\text{P}^{d}_{\text{ref}}$, in per-unit) in the reference case from
PGLib, and the range of total active power demand in the dataset ($\underline{\text{P}}^{d}, \overline{\text{P}}^{d}$, see details below).

\begin{table}[!t]
    \centering
    \caption{Test Systems Statistics}
    \label{tab:results:pglib}
    \begin{tabular}{lrrrrr}
    \toprule
        System 
        & \multicolumn{1}{c}{$|\mathcal{N}|$}
        & \multicolumn{1}{c}{$|\mathcal{E}|$}
        & \multicolumn{1}{c}{\#indep. }
        & \multicolumn{1}{c}{${\text{P}^{d}_{\text{ref}}}^{\dagger}$}
        & \multicolumn{1}{c}{$[\underline{\text{P}}^{d}, \overline{\text{P}}^{d}]^{\ddagger}$} \\
    \midrule
        \ieeeXXS
            & 14
            & 20
            & 168
            & 2.6
            & [\phantom{000}1.9, \phantom{000}2.9]
            \\
        \ieeeXS 
            & 118
            & 186
            & 1,538
            & 42.4
            & [\phantom{00}33.4, \phantom{00}51.8]
            \\
        \ieeeS 
            & 300
            & 411
            & 3,477
            & 235.3
            & [\phantom{0}184.8, \phantom{0}250.3]
            \\
        \pegaseSmall 
            & 1354
            & 1991
            & 16,645
            & 730.6
            & [\phantom{0}581.3, \phantom{0}772.3]
            \\
        \pegaseMedium   
            & 2869
            & 4585
            & 37,833
            & 1324.4
            & [1053.9, 1529.8]
            \\
    \bottomrule
    \end{tabular}\\
    {\footnotesize $^{\dagger}$Reference total active power load. $^{\ddagger}$Range of total load in the dataset.}
\end{table}

For each system, instances are generated by perturbing loads in the same fashion as \cite{Mak2023_LearningACOPF_ADMM,Chen2023_E2ELR}.
Namely, the active and reactive loads $\pd, \qd \, {\in} \, \mathbb{R}^{N}$ are sampled as
\begin{align}
    \pd = \alpha \cdot \eta \circ \pdRef, \quad \qd = \alpha \cdot \eta \circ \qdRef
\end{align}
where $\pdRef, \qdRef \, {\in} \, \mathbb{R}^{|\mathcal{N}|}$ are the vectors of nominal active and reactive demand in the PGLib test case, $\alpha \, {\in} \, \mathbb{R}$ is a system-wide scaling factor, $\eta \, {\in} \, \mathbb{R}^{|\mathcal{N}|}$ is a vector of uncorrelated, multiplicative noise, and $\circ$ denotes element-wise product.
The system-wide scaling factor $\alpha$ is sampled from a uniform distribution $U[l, u]$ where $l {=} 0.8$ and $u$ ranges from $1.05$ to $1.20$ depending on system capacity, and $\eta$ is sampled from a log-normal distribution $\text{LogNormal}(\frac{-\sigma^{2}}{2}, \sigma^{2})$  with $\sigma \, {=} \, 0.05$.
The last column of Table \ref{tab:results:pglib} reports the range of total active power demand in the dataset generated.

\newcommand{\clarabelQuad}{Clarabel\textsubscript{quad}}

For each system, a dataset of SOC-OPF instances and their solutions is generated.
Each instance is formulated using PowerModels \cite{power_models} and solved
\revision{twice: once with Mosek 10 \cite{mosek}, and once with Clarabel v0.7 \cite{clarabel} in quadruple (128-bit) floating-point arithmetic, denoted by \clarabelQuad{} in what follows.
Mosek uses tolerances of $10^{-6}$, in line with \cite{Bienstock2024_AccurateSOCP}, whereas \clarabelQuad{} uses tolerances of $10^{-12}$ to obtain high-precision solutions.
The use of extended precision alleviates numerical issues typically encountered by conic interior-point solvers when solving SOC-OPF problems \cite{Bienstock2024_AccurateSOCP}, thus allowing to obtain very high-quality solutions, albeit at the cost of increased computing times.
In all that follows, the solutions obtained by \clarabelQuad{} are considered as ``ground-truth" primal-dual optima.
Both solvers use a single CPU core to facilitate solving all instances in parallel.
Gurobi was not used in the experiments because its API does not support querying conic dual variables.}

\revision{For each system, a total of 65,536 instances are generated. Mosek failed to converge for 1690, 0, 3269, 72 and 0 instances for the \ieeeXXS, \ieeeXS, \ieeeS, \pegaseSmall{} and \pegaseMedium{} systems, respectively: specifically, JuMP reported the termination status of these instances to be \texttt{SLOW\_PROGRESS}, and feasible primal-dual pairs were not found.
\clarabelQuad{} did not report any numerical issue.
Out of these instances, \clarabelQuad{} finds 2087, 0, 3890, 221 and 1 instances to be infeasible for the systems respectively, and finds solutions for all remaining instances.
A dataset of 30,000 instances is then created by randomly selecting instances for which a primal-dual pair was found by both solvers.}
Finally, each dataset is split into training (90\%), validation (5\%) and testing (5\%).
The solutions returned by \revision{Mosek and \clarabelQuad{}} are only used for evaluating optimality gaps during testing, i.e., they are not used in training or validation.

The DCP proxies are implemented in Python 3.10 using PyTorch 2.0 \cite{paszke2019pytorch} with the Adam optimizer \cite{kingma2014adam}.
All experiments are conducted on Intel Xeon 2.7GHz CPU machines running Linux and equipped with Tesla RTX 6000 GPUs, at the Phoenix cluster \cite{PACE}.

\subsection{\revision{Performance Evaluation}}
\label{sec:results:Mosek}

\revision{Given the numerical challenges of solving SOC-OPF problems, it is reasonable to expect that the dual solution returned by Mosek may not satisfy all (dual) constraints to high accuracy.
Therefore, the dual bound returned by the solver may be invalid.}
\revision{Such behavior has been reported in \cite{Bienstock2024_AccurateSOCP}, and in} \cite{Oustry2022_CertifiedSDPBounds} in the context of SDP relaxations of AC-OPF.

\revision{To measure this effect, Table \ref{tab:results:violation} reports, for each system, the maximum constraint violation of Mosek's dual solutions in the testing set.}
Mosek uses an interior-point algorithm that strictly enforces variable bounds and conic constraints \eqref{eq:DSOC:non_negative}--\eqref{eq:DSOC:cone:omega};
constraint violations therefore occur on \revision{dual} equality constraints, especially \eqref{eq:DSOC:wm}--\eqref{eq:DSOC:wi}.
\revision{Recall that, in DSOC-OPF, the equality constraints \eqref{eq:DSOC:pf}--\eqref{eq:DSOC:wi} all have zero right-hand side, thus, only absolute violations are meaningful.}

\revision{The quality of the DCP and Mosek dual solutions is evaluated in terms of optimality gap with respect to the objective value obtained by \clarabelQuad{}.
Recall that the DCP solutions are always dual-feasible, hence the associated dual bound is always valid.
To ensure a fair comparison, the same dual recovery steps used in the DCP proxies are applied to repair the Mosek solutions.
Note that different dual recovery steps may yield different dual bounds for the same Mosek dual solution,
The optimality gap is defined as 
\begin{align*}
    \text{gap} = \frac{z^{*} - \hat{z}}{|z^{*}|},
\end{align*}
where $z^{*}$ is the dual objective value of the \clarabelQuad{} solution, and $\hat{z}$ denotes the objective value of a candidate dual solution, obtained from DCP or Mosek (after dual recovery).
Finally, the paper reports averages using the geometric mean $\mu(x_{1}, ..., x_{n}) \, {=} \sqrt[n]{x_{1} x_{2} ... x_{n}}$.}

    \begin{table}[!t]
        \centering
        \caption{Maximum Constraint Violation of Mosek Dual Solutions.}
        \label{tab:results:violation}
        \color{revColor}
        \begin{tabular}{lrrrrrrrrr}
            \toprule
            System
                & \eqref{eq:DSOC:pg}--\eqref{eq:DSOC:qg}
                & \eqref{eq:DSOC:pf}--\eqref{eq:DSOC:qt}
                & \eqref{eq:DSOC:wm}
                & \eqref{eq:DSOC:wr}
                & \eqref{eq:DSOC:wi}
                \\
            \midrule
            \ieeeXXS 
                & 0.00
                & 0.00
                & 0.02
                & 0.02
                & 0.02 
                \\
            \ieeeXS
                & 0.01
                & 0.00
                & 0.95
                & 0.17
                & 1.23
                \\
            \ieeeS 
                & 0.08
                & 0.10 
                & 516.51
                & 73.74
                & 211.61
                \\
            \pegaseSmall 
                & 0.01
                & 0.01
                & 93.24 
                & 75.62 
                & 12.16
                \\
            \pegaseMedium 
                & 0.01
                & 0.01
                & 100.48
                & 100.77
                & 25.64
                \\
            \bottomrule
        \end{tabular}\\
        \smallskip
        {\footnotesize Constraints \eqref{eq:DSOC:non_negative}--\eqref{eq:DSOC:cone:omega} are always satisfied. All dual violations in per-unit.}
    \end{table}

\subsection{Dual Conic Proxy Performance}
\label{sec:results:DCP}

The experiments consider different realizations of the DCP
architecture on multiple test case systems, i.e., it considers
different sets of independent variables and the use of polar vs. rectangular
form as outlined in Section
\ref{sec:architecture:rect_vs_polar}.  Since the
magnitude of different types of dual variables varies greatly in scale, the experiments
also study the impact of scaling the outputs of different sub-networks \revision{by empirically chosen scaling factors that are in powers of 10}. Since this scaling
is found to improve optimality gaps \revision{in most cases}, only these results are
shown. Due to space limitations, only the following approaches are
reported: (1) whether to predict $\lambdaPf$, $\lambdaQf$, $\lambdaPt$, $\lambdaQt$ and recover $\nuThermalPfr$, $\nuThermalQfr$, $\nuThermalPto$, $\nuThermalQto$, or vice versa, and (2) whether to predict
$(\omegat, \omegaf)$ in rectangular form or polar form.

\begin{table}[!t]
    \centering
    \caption{\revision{DCP Performance Results}}
    \label{tab:results:gaps}
    \color{revColor}
    \resizebox{\columnwidth}{!}{
    \begin{tabular}{lccrrrrrr}
        \toprule
            &&
            & \multicolumn{3}{c}{\%gap -- DCP}
            & \multicolumn{3}{c}{\%gap -- Mosek$^{\dagger}$}
        \\
        \cmidrule(lr){4-6}
        \cmidrule(lr){7-9}
            System
            & \multicolumn{1}{c}{indep.}
            & \multicolumn{1}{c}{$(\omegaf, \omegat)$}
            % DCP gaps
            & \multicolumn{1}{c}{mean}
            & \multicolumn{1}{c}{std}
            & \multicolumn{1}{c}{max}
            % Mosek gaps
            & \multicolumn{1}{c}{mean}
            & \multicolumn{1}{c}{std}
            & \multicolumn{1}{c}{max}
        \\
\midrule
\ieeeXXS
    & $\lambdaf, \lambdat$ & Rect.
        & 1.17
        & 5.50
        & 25.20
        & 0.00
        & 0.00
        & 0.01
        \\
    & $\lambdaf, \lambdat$ & Polar
        & 0.32
        & 5.53
        & 24.79
        & 0.00
        & 0.00
        & 0.00
        \\
    & $\nuThermalfr, \nuThermalto$ & Rect.
        & 0.54
        & 5.48
        & 24.68
        & 0.00
        & 0.00
        & 0.01
        \\
    & $\nuThermalfr, \nuThermalto$ & Polar
        & \textbf{0.05}
        & 5.51
        & \textbf{24.52}
        & 0.00
        & 0.00
        & 0.01
        \\
\midrule
\ieeeXS
    & $\lambdaf, \lambdat$ & Rect.
        & 0.76
        & 0.49
        & 2.51
        & 0.00
        & 0.00
        & 0.00
        \\
    & $\lambdaf, \lambdat$ & Polar
        & 0.40
        & 0.45
        & 2.07
        & 0.00
        & 0.00
        & 0.00
        \\
    & $\nuThermalfr, \nuThermalto$ & Rect.
        & 0.36
        & 0.09
        & 0.98
        & 0.00
        & 0.00
        & 0.00
        \\
    & $\nuThermalfr, \nuThermalto$ & Polar
        & \textbf{0.08}
        & 0.15
        & \textbf{0.57}
        & 0.00
        & 0.00
        & 0.00
        \\
\midrule
\ieeeS
    & $\lambdaf, \lambdat$ & Rect.
        & 1.47
        & 1.46
        & 14.86
        & 0.00
        & 0.01
        & 0.18
        \\
    & $\lambdaf, \lambdat$ & Polar
        & 0.94
        & 1.46
        & 14.45
        & 0.00
        & 0.01
        & 0.18
        \\
    & $\nuThermalfr, \nuThermalto$ & Rect.
        & 0.86
        & 1.37
        & 14.11
        & 0.00
        & 0.01
        & 0.18
        \\
    & $\nuThermalfr, \nuThermalto$ & Polar
        & \textbf{0.58}
        & 1.27
        & \textbf{13.56}
        & 0.00
        & 0.01
        & 0.18
        \\
\midrule
\pegaseSmall
    & $\lambdaf, \lambdat$ & Rect.
        & 54.86
        & 5.01
        & 63.28
        & 0.02
        & 0.01
        & 0.06
        \\
    & $\lambdaf, \lambdat$ & Polar
        & 54.77
        & 5.02
        & 63.20
        & 0.02
        & 0.01
        & 0.06
        \\
    & $\nuThermalfr, \nuThermalto$ & Rect.
        & 2.14
        & 0.23
        & 2.66
        & 0.01
        & 0.00
        & 0.04
        \\
    & $\nuThermalfr, \nuThermalto$ & Polar
        & \textbf{1.62}
        & 0.18
        & \textbf{2.08}
        & 0.01
        & 0.00
        & 0.04
        \\
\midrule
\pegaseMedium
    & $\lambdaf, \lambdat$ & Rect.
        & 67.82
        & 4.86
        & 75.28
        & 0.02
        & 0.01
        & 0.05
        \\
    & $\lambdaf, \lambdat$ & Polar
        & 67.53
        & 4.90
        & 75.06
        & 0.02
        & 0.01
        & 0.05
        \\
    & $\nuThermalfr, \nuThermalto$ & Rect.
        & 2.09
        & 0.17
        & 2.62
        & 0.02
        & 0.00
        & 0.04
        \\
    & $\nuThermalfr, \nuThermalto$ & Polar
        & \textbf{1.57}
        & 0.18
        & \textbf{2.11}
        & 0.02
        & 0.00
        & 0.04
        \\
        \bottomrule
    \end{tabular}}\\
    \smallskip
    \footnotesize{All gaps are w.r.t. \clarabelQuad{}. $^{\dagger}$After dual feasibility restoration.}
\end{table}

\begin{table}[!t]
    \centering
    \caption{\revision{Geometric Means of Computing Time of DCP and Solvers}}
    \label{tab:results:time}
    \color{revColor}
    \begin{tabular}{l rrr}
        \toprule
         \multicolumn{1}{c}{System} & DCP & Mosek & \clarabelQuad \\
     \midrule
       \ieeeXXS & 4\,ms & 0.01\,s & 0.12\,s \\
       \ieeeXS & 21\,ms & 0.44\,s & 1.32\,s \\
       \ieeeS & 15\,ms & 1.26\,s & 5.28\,s \\
       \pegaseSmall & 32\,ms & 10.19\,s & 33.14\,s \\
       \pegaseMedium & 66\,ms & 51.32\,s & 90.32\,s \\
       \bottomrule
    \end{tabular}
    \\
    \smallskip
    {\footnotesize Mosek and \clarabelQuad{} times are per instance, using one CPU core.\\
    DCP times are per batch of 512 instances, using a GPU.}
\end{table}

\revision{Table \ref{tab:results:gaps} reports, for each system and DCP architecture, the mean, standard deviation, and maximum optimality gaps
achieved by DCP and by Mosek.
Recall that all gaps are measured against the optimal value obtained by \clarabelQuad{}. 
The second and third columns indentify the DCP architecture, namely, whether $\lambdaf, \lambdat$ or $\nuThermalfr, \nuThermalto$ are selected as
independent variables, and whether $(\omegaf, \omegat)$ is represented using rectangular or polar form.
All optimality gaps for Mosek are computed from the repaired Mosek solutions, using the same recovery steps as the DCP architecture.
}

\revision{The results in Table \ref{tab:results:gaps} highlight some important points.} First, selecting $\nuThermalfr, \nuThermalto$ as independent
variables always outperforms $\lambdaf, \lambdat$.  This is
potentially because most branches are never congested in any instance,
in which case $\nuThermalfr \, {=} \, \nuThermalto \, {=} \, 0$, which
greatly simplifies the learning task.  In addition, the polar
representation of $(\omegaf, \omegat)$ also outperforms the
rectangular representation.  As mentioned in Section
\ref{sec:architecture:rect_vs_polar}, this may be explained by the
fact that the distribution of angles is typically centred around
$\pi/4$ and exhibits very small variance, which also simplifies the
learning task.

\revision{Second, although Mosek's dual solutions may exhibit non-trivial violations, passing them through the dual completion steps yields overall high-quality dual bounds.
The large maximum optimality gap of 0.18\%, observed on the \ieeeS{} system, is mostly likely caused by numerical difficulties, and emphasizes the importance of ensuring valid dual bounds.
Furthermore, the original (possibly invalid) dual bound returned by Mosek, i.e., before repairing dual feasibility, is on average within 0.03\% of the optimal value; this is compared to average gaps of at most 0.02\% after feasibility recovery.
%Therefore, applying the proposed dual feasibility recovery step increases, on average, Mosek's dual objective value.
{\em These findings highlight the broader value of the proposed dual completion procedure: it can be used in conjunction with an optimization solver, to obtain high-quality valid dual bounds.}}

\revision{Third, the best DCP proxies provide dual solutions with valid dual bounds that achieve average gaps below $0.1\%$ for the smaller systems (\ieeeXXS, \ieeeXS), about 0.5\% on the medium-size \ieeeS, and about 1.6\% for the larger \pegaseSmall{} and \pegaseMedium{} systems.
One striking observation on the larger Pegase systems is that DCP also achieves good worst-case optimality gaps, of about 2\% gap.
This is especially important, because this worst-case performance is only 25\% higher than the average performance, in contrast to that of the smaller IEEE systems where the worst gap may be 500x larger than the average gap.}

\revision{Finally, Table \ref{tab:results:time} reports mean computing times for DCP, Mosek and \clarabelQuad{}.
The computing time of DCP is reported per batch of 512 instances on a GPU under the same conditions as previously described.
The computing time for Mosek and \clarabelQuad{} are per instance, using a single CPU core.
Computing times for Mosek and \clarabelQuad{} were recorded during the data-generation process, during which 24 instances were solved simultaneously on a single CPU node, which inevitably yields higher computing times because of limited thread availability (one core per instance), cache access, and potential CPU frequency slowdown because of high CPU usage.}
\revision{Nevertheless,} the proxies significantly outperform Mosek in terms of
computing times, especially for larger instances.  For
\pegaseMedium{}, using the best DCP configuration, evaluating a batch of
512 instances takes \revision{an average 66ms} on a GPU, whereas solving one instance takes
\revision{51.32s} on a single CPU core.
\revision{{\em Overall, the DCP proxies yield three orders of magnitude} speedups over Mosek. These speedups allow DCP to provide, for the first time, performance guarantees for AC-OPF proxies in real time.}

\section{Conclusion}
\label{sec:conclusion}

The paper has proposed, for the first time, dual optimization proxies
that provide valid high-quality lower bounds for AC-OPF in milliseconds.  The
proposed dual conic proxies build on convex relaxations of AC-OPF and
conic duality.  A core contribution of the paper is the proposed
dual feasibility completion that guarantees dual feasibility, thus
providing valid dual bounds by leveraging fundamental properties of
dual-optimal solutions.  This dual feasibility completion is not
restricted to ML settings: it can provide certified dual bounds given
close-to-feasible solutions produced by an optimization solver.
Extensive numerical experiments on power grids with up to 2869 buses
have demonstrated that such proxies can produce high-quality bounds
within very short time, improving computing times by three orders of
magnitude compared to a state-of-the-art optimization solver.
Relevant directions of future work include the design of enhanced
completion methods to further improve proxy performance, and applying
the DCP framework to DC-OPF and other convex relaxations of AC-OPF
such as the SDP relaxation.

\bibliographystyle{IEEEtran}
\bibliography{ref}

\end{document}